\documentclass[letterpaper, 10 pt, journal, twoside]{IEEEtran}
\ifCLASSINFOpdf
\else
\fi

\usepackage{graphics} 
\usepackage{epsfig} 
\usepackage{mathptmx} 
\usepackage{times} 
\usepackage{amsmath} 
\usepackage{bm,amsmath}
\usepackage{amssymb}  
\usepackage{multirow}
\usepackage{color,soul}
\usepackage{algorithm}
\usepackage{siunitx}
\usepackage{algpseudocode}
\usepackage{xcolor}

\usepackage{url}

\begin{document}
%
\title{Mixed Reality Teleoperation Assistance for Direct Control of Humanoids}
%
%
%

\author{Luigi Penco$^{1}$, Kazuhiko Momose$^{2}$, Stephen McCrory$^{1,3}$, Dexton Anderson$^{1}$, Nicholas Kitchel$^{1}$, \\ 
Duncan Calvert$^{1,3}$, Robert J. Griffin$^{1,3}$%
\thanks{Manuscript received: August, 14, 2023; Revised October, 12, 2023; Accepted December, 17, 2023.}
\thanks{This paper was recommended for publication by Editor Jee-Hwan Ryu upon evaluation of the Associate Editor and Reviewers' comments.
This work was supported by ONR Grant N00014-22-1-2593 and DAC Cooperative Agreement W911NF-22-2-0201.} 
\thanks{$^{1}$The authors are with the Florida Institute for Human and Machine Cognition, 40 S Alcaniz St, Pensacola, FL 32502, United States
        {\tt\footnotesize lpenco@ihmc.org}}%
\thanks{$^{2} $The author is with the Florida Institute of Technology, 150 W University Blvd, Melbourne, FL 32901, United States
        {\tt\footnotesize kmomose@my.fit.edu}}%
\thanks{$^{3} $The authors are with the University of West Florida, 11000 University Pkwy, Pensacola, FL 32514, United States
        {\tt\footnotesize rgriffin@uwf.edu}}
\thanks{Digital Object Identifier (DOI): see top of this page.}
}
%
%

\markboth{IEEE Robotics and Automation Letters. ArXiv Version. Accepted December, 2023}
{Penco \MakeLowercase{\textit{et al.}}: Mixed Reality Teleoperation Assistance for Direct Control of Humanoids} 

%



\maketitle

\begin{abstract}
Teleoperation plays a crucial role in enabling robot operations in challenging environments, yet existing limitations in effectiveness and accuracy necessitate the development of innovative strategies for improving teleoperated tasks. This article introduces a novel approach that utilizes mixed reality and assistive autonomy to enhance the efficiency and precision of humanoid robot teleoperation. By leveraging Probabilistic Movement Primitives, object detection, and Affordance Templates, the assistance combines user motion with autonomous capabilities, achieving task efficiency while maintaining human-like robot motion. Experiments and feasibility studies on the Nadia robot confirm the effectiveness of the proposed framework. Supplementary video available at \url{https://youtu.be/oN-FD6YnF2c}.
\end{abstract}


%
\IEEEpeerreviewmaketitle

\section{Introduction}
%
%
%
%
\IEEEPARstart{S}{everal} of the most compelling instances of humanoids performing valuable tasks have been through teleoperation \cite{kourosh2021survey}. However, despite years of research in this domain, teleoperated humanoid robots continue to show substantial limitations when executing tasks under user control.
Traditional direct control of robots through the user's own motions offers an intuitive approach for interacting with the remote environment \cite{ishiguro2018,penco2019,kourosh2021survey}. However, differences in kinematics between humans and robots can lead to errors and multiple attempts for task completion, making this approach not very effective and time-consuming, and even lead to operator frustration.

\begin{figure}[!t]
\centering
\includegraphics[width=\linewidth]{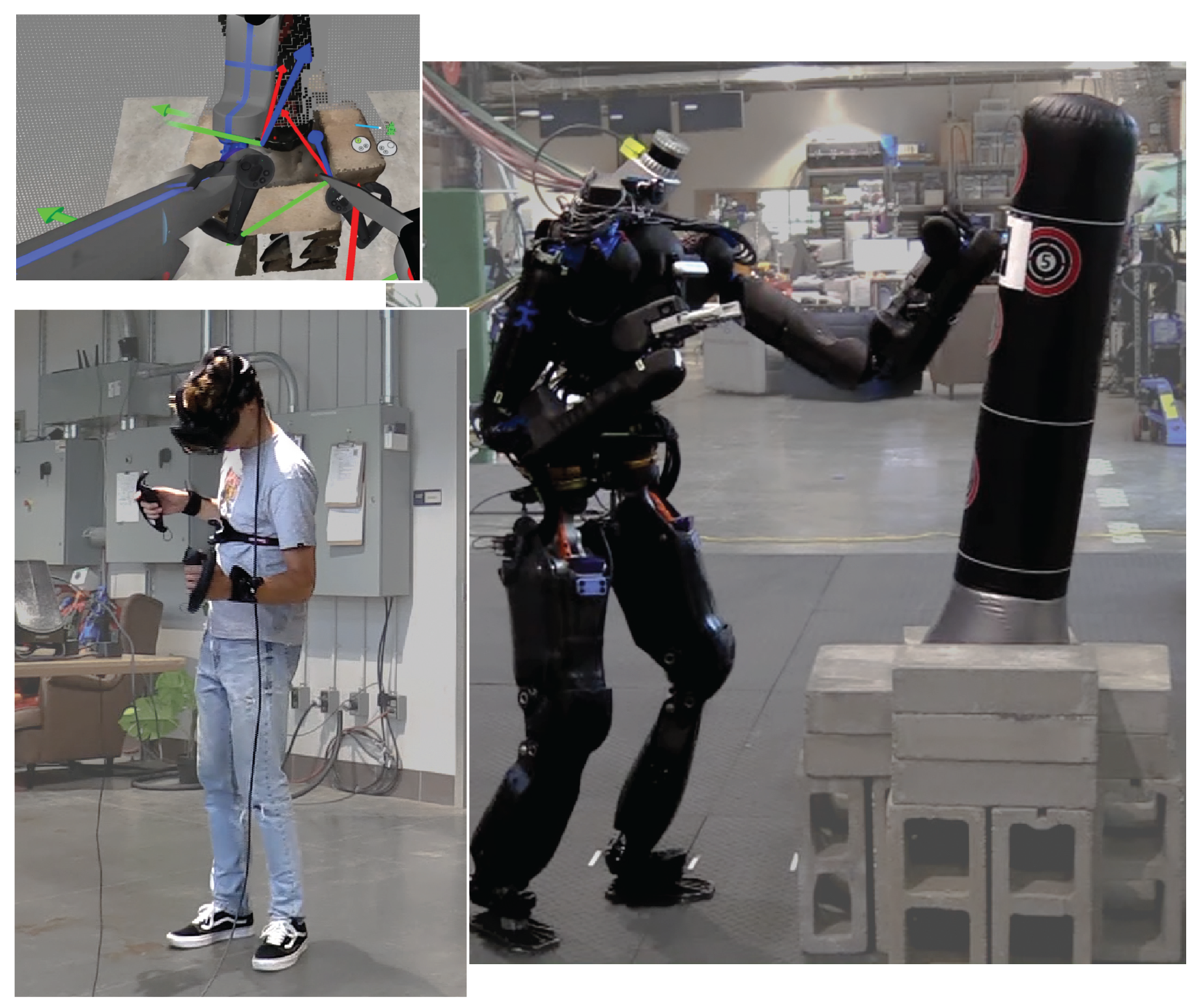}
\caption{With the teleoperation assistance we propose, the user serves as a high-level guide, directing the robot's autonomous task execution through a mixed reality interface (top left) and specifying the manner in which the task should be carried out. In the given example, the user merely initiates a portion of a punching motion, specifically indicating an uppercut technique. Recognizing this, the robot discerns that it is tasked with a punching action and, among the various techniques available, it should execute an uppercut. The user does not need to focus on the precision of the subsequent motion, as the initial input is sufficient for the robot to complete the task accurately.}
\label{fig:intro}
\end{figure}

Our work addresses the challenge of improving the direct control of humanoids in teleoperation by introducing autonomous assistance features in mixed reality. The goal is to enhance the efficiency and accuracy of teleoperated tasks. To achieve this, we employ Probabilistic Movement Primitives (ProMPs) \cite{paraschos2017}, which are widely used for modular movement representation and generation. These ProMPs are conditioned on the operator's input, allowing adaptation to the user's preferred way of performing the task while maintaining similarity to the demonstrations. Additionally, we utilize information from the robot's sensors to provide context and precision to the motion primitives.
The physical interaction with objects is governed by Affordance Templates (ATs) \cite{pettinger2020}, ensuring accurate and controlled manipulation while reducing the risk of errors or unintended consequences during physical interaction.
In order to observe, predict and direct the assistance, we developed a mixed reality interface, based on Coactive Design principles \cite{johnson2014coactive}. Using mixed reality has the benefit of providing an immersive experience for the operator as also shown in \cite{cockpitJorgensen}.

The need of such teleoperation assistance was vividly highlighted during the finals of the ANA Avatar XPRIZE \cite{vaz2022,luo2022,schwarz2023robust}, a four-year global competition on robot teleoperation that ended in November 2022. Operators struggled to effectively and accurately control the robots, even for seemingly simple tasks like grasping a bottle. Multiple attempts were often required, showcasing the difficulties in achieving satisfactory outcomes solely through direct user control — especially when considering that robots can perform similar tasks autonomously with relative ease.

It is important to note that full direct control is typically achieved by projecting the RGB images from the robot camera to the Virtual Reality (VR) headset \cite{luo2022,schwarz2023robust}, but this approach often fails to provide sufficient situational awareness \cite{endsley1995toward} to the user. This was also demonstrated in the ANA Avatar XPRIZE finals, where judge operators frequently encountered confusion regarding the robot's location within the environment. Hence, a combination of autonomy and mixed reality technology is crucial in bridging these gaps and enabling robots for real-world applications. 

\begin{figure*}[t!]
\centering
\includegraphics[width=\linewidth]{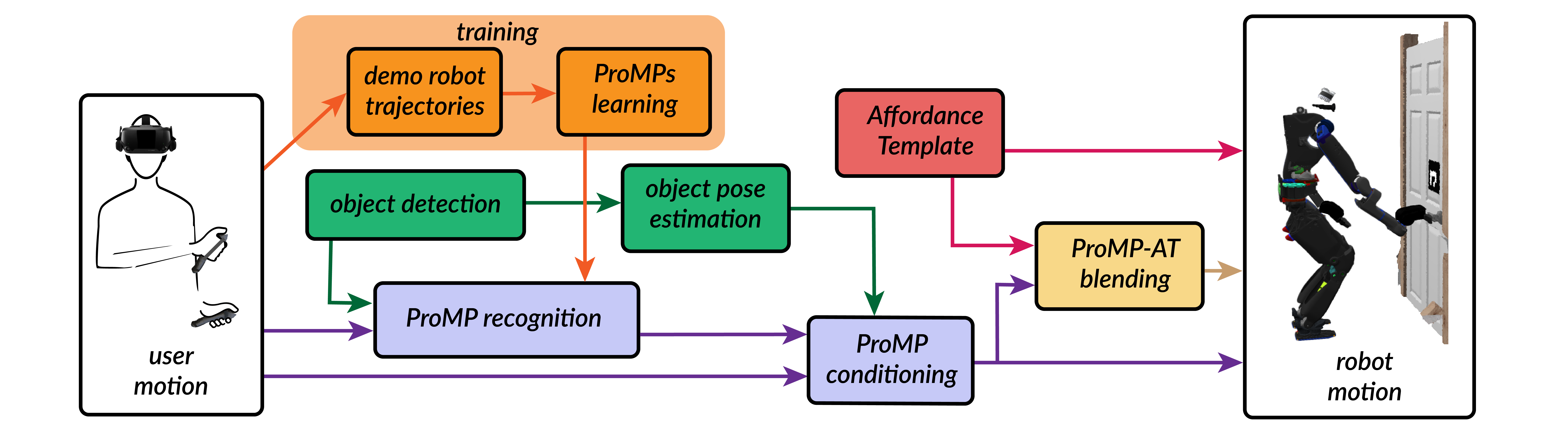}
\caption{Flowchart of the assistive autonomy. During the training phase, the human
operator teleoperates the robot in simulation, and performs the tasks in different ways.  A ProMP is learned for every task.  When teleoperating the robot in real-time, the ProMPs are used to generate the assistive robot motion: (1) the system
recognizes the current task, with the help of context given by the object detection; (2) it updates the ProMP according to the initially observed user input and object pose; (3) if an affordance is available for that object, a blending mechanism is used to to achieve smooth transitions between ProMP-generated motions and ATs.}
\label{fig:flowchart}
\end{figure*}

A recent and promising approach to addressing the direct control issues is shared-control, which integrates robot assistance with user input to facilitate the completion of tasks \cite{dragan2013a}. For instance, Rakita \textit{et al.} \cite{rakita2019} implemented a shared-control approach in which they teleoperated the upper body of a humanoid robot. By maintaining a constant offset between the end effectors or limiting the motion of the end-effector that holds the object, the system provides on-the-fly assistance and helps the user perform bimanual tasks more easily. Similarly, in \cite{rahal2019},  Rahal \textit{et al.} designed a shared control approach to assist the human operator by enforcing different nonholonomic-like constraints representative of the cutting kinematics.

In many shared-control frameworks, the user provides an input that allows the robot to consider the human's intent and to assist in the task by adjusting its motion or executing a pre-optimized version of that motion. To determine the final reference for the robot's motion, a policy blending mechanism that governs the balance between the user input and the enhanced robot motion  is employed  \cite{dragan2013a}. The blending policy is usually the distance to the goal. The closer the robot gets to a predicted goal, the more likely it is that the predicted goal is the correct one, and thus, the robot input is favored over the user's. 

Approaches have also been proposed that consider unknown goals. For instance, in the case of \cite{javdani2018}, a POMDP-based strategy has been employed to provide assistance, aiming to minimize the expected cost-to-go. It is worth highlighting that this particular approach is predominantly designed for the control of goal-oriented components of the robot, such as its end-effectors. Consequently, its direct applicability to the control of other robot components, such as adjusting the orientation of the chest and forearms as needed for whole-body tasks may be limited.
Alternatively, \cite{broad2020} presents a novel approach leveraging the Koopman operator for learning the dynamics of a machine and enhancing human-machine interaction. However, the application and scalability of this system to more complex systems, such as humanoid robots, remain unvalidated and appear to be challenging due to the increased complexity and unpredictability inherent in such systems.

In haptic-based shared-control approaches, a different strategy is employed. Here, only the user's input actuates the robot, but haptic information is used to guide the user during task execution \cite{rahal2019,selvaggio2022}. By incorporating haptic feedback, the system can provide the user with tactile cues or force feedback, improving the user's understanding of the task and making the teleoperation process more intuitive and effective.

In line with the efforts that have been made in humanoid teleoperation, our work continues to advance these endeavors. 
By integrating predictive modeling techniques, we are poised to proactively gauge the user's intended motion, tailoring the robot behavior to both fit the user input and the pose of manipulated objects in the environment.
By leveraging the strengths of both the user and the robot, we do not just enhance task efficiency; we also maintain the human-likeness of the robot's motion.
Our work is validated through experiments and feasibility studies on the Nadia humanoid robot.

\section{Assistive Autonomy}
\label{sec:assistance}
Our assistive autonomy seeks to improve robot teleoperation by enabling more precise and adaptive robot movements. This approach ensures fidelity to the demonstrated actions while compensating for inaccuracies in user input, both during the demonstrations and during actual operation. The proposed method utilizes ProMPs along with ATs when physical interaction with the object is required.

The system follows a series of operations as shown in Fig. \ref{fig:flowchart}. The learning phase involves recording few demonstrations (less than 30) of the task performed in different ways by a human operator in a simulated environment. ProMPs are learned from these demonstrations. During teleoperation, the system recognizes the current task by identifying the corresponding ProMP, aided by semantic object information. The system then updates the posterior distribution of the ProMPs using the initial user input and the estimated object pose. The mean trajectory of the updated ProMP is used as a reference for the robot controller.
If an affordance is available for that object, a blending mechanism is used to achieve smooth transitions between ProMP-generated motions and ATs. 
The generated trajectories are then tracked by the robot controller (Section \ref{sec:control}), which calculates the appropriate joint commands. Both the object pose conditioning and the AT integration make it possible to transfer the learned ProMPs to the real world without any real-world training data.

\subsection{Probabilistic Movement Primitives (ProMPs)}
ProMPs \cite{paraschos2017} are a probabilistic model used to represent trajectory distributions. These models describe the time-varying mean and variance of trajectories using basis functions. Each trajectory is parameterized by a weight vector $\bm{w}\in \mathbb{R}^m$.  
Given the weight vector, the probability $\bm{\xi}(t)$ of observing a trajectory $\bm{y}$ is modeled using a linear basis function model:
\begin{align}
&\bm{\xi}_y(t) = \bm{\Phi}_t\bm{w}+\bm{\epsilon}_\xi,
\label{eq:promp}
\end{align}
where $\bm{\Sigma}_\xi$ represents the observation noise variance, and $\bm{\epsilon}_\xi\sim\mathcal{N}(0,\bm{\Sigma}_\xi)$ is the trajectory noise. $\bm{\Phi}_t \in \mathbb{R}^{n \cdot m}$ is a block-diagonal matrix containing the $m$ basis functions for each dimension $n$.
The distribution $p(\bm{w};\bm{\theta})$ over the weight vector $\bm{w}$ is Gaussian, with parameters $\bm{\theta}={\bm{\mu_w},\bm{\Sigma_w}}$ specifying the mean and variance of $\bm{w}$.

\subsection{Learning ProMPs from Demonstrations}
\label{sec:learn}
In the learning phase of ProMPs, demonstrations are recorded from a human operator teleoperating the robot in simulation to perform the task in different ways. To decouple the movement from the time signal and obtain a duration-independent representation, a phase variable $\upsilon \in [0,1]$ is introduced. The modulated trajectories $\bm{\xi}_i(\upsilon)$ are then used to learn a ProMP for the task. The parameters $\bm{\theta}={\bm{\mu}_w,\bm{\Sigma}_w}$ of the ProMP are estimated using a maximum likelihood estimation algorithm. The weight vectors $\bm{w}_i$ for each demonstration $i$ are computed using linear ridge regression, as shown in (\ref{eq:promp3}), where the ridge factor $\lambda$ is typically set to a very small value ($\lambda=10^{-12}$ in our case):
\begin{equation}
\bm{w}_i=\big(\bm{\Phi}_\upsilon^\top\bm{\Phi}_\upsilon+\lambda\big)^{-1}\bm{\Phi}_\upsilon^\top\bm{\xi}_i(\upsilon),
\label{eq:promp3}
\end{equation}

Assuming normal distributions $p(\bm{w})\sim\mathcal{N}(\bm{\mu}_w,\bm{\Sigma}_w)$, the mean $\bm{\mu}_w$ and covariance $\bm{\Sigma}_w$ can be computed from the samples $\bm{w}_i$:
\begin{equation}
\bm{\mu_w}=\frac{1}{D}\sum_{i=1}^{D} \bm{w}_i, ~~~~\bm{\Sigma_w}=\frac{1}{D}\sum_{i=1}^{D} (\bm{w}_i-\bm{\mu_w})(\bm{w}_i-\bm{\mu_w})^\top,
\label{eq:meanvariance}
\end{equation}
where $D$ is the number of demonstrations.
Each ProMP models the behavior of an individual body part of the robot and is learned in the object frame. This means that the movements are represented and learned relative to the object or task being performed, rather than being tied to a specific robot or coordinate system. This decoupling of the movement from the robot's frame of reference allows for greater flexibility in executing the learned movements in different environments. 


\subsection{Recognizing the Motion Primitive}
\label{sec:recognition}
Given the current context (detected object), we determine to which ProMP the present teleoperated motion belongs by minimizing the distance between the initial $n_{obs}$ observations (typically corresponding to a fourth or a third of the duration of a given motion) and the ProMP's mean:
\begin{equation}
\hat{k}= \text{arg}\min_{k \in [1:K]}\bigg[ \sum_{t\in T_{obs}}\| \bm{y}(t)-\Phi_{t}\bm{\mu}_{\bm{w}_{k}}\| \bigg],
\label{eq:distance}
\end{equation}
where $K$ is the number of tasks in the dataset associated to the detected object (Section \ref{sec:object}) and $T_{obs}=\{t_1,...,t_{n_{obs}}\}$ is the set of timesteps associated to the $n_{obs}$ early observations. 
While computing $\hat{k}$, the ProMP is modulated to have a duration equal to the mean duration of the demonstrations.
The recognition (\ref{eq:distance}) starts whenever a motion is detected, i.e. the derivative of the observed end-effector trajectories exceeds a given threshold, after the user activates the assistance mode (Section \ref{sec:interface}). This ensures that motion data is not processed as observations used for conditioning in instances where the user remains inactive.

\subsection{ProMPs Conditioning}
\label{sec:conditioning}
Once the right ProMP has been identified, we update their posterior distribution to take into account the initial observations from the user input.
Conditioning is beneficial for tailoring the motion to accommodate the user's preferred method of task execution. If this step is omitted, the system would default to executing the average value of the learned data, which may not align with the user's intentions.
Each ProMP has to be conditioned to reach a certain observed state $\bm{y}^*_{t}$.
The conditioning for a given observation $\bm{x}^*_{t}=\{\bm{y}^*_{t},\bm{\Sigma}^*_{y}\}$ (with $\bm{\Sigma}^*_{y}$ being the noise in the desired observation) is performed by applying Bayes' theorem:
\begin{equation}
p({\bm{w}_{\hat{k}}}|\bm{x}^*_{t})\propto \mathcal{N}(\bm{y}^*_{t}|\bm{\Phi}_{t}{\bm{w}_{\hat{k}}},\bm{\Sigma}^*_y)p({\bm{w}_{\hat{k}}}).
\label{eq:bayes}
\end{equation}
The conditional distribution of $p({\bm{w}_{\hat{k}}}|\bm{x}^*_{t})$ is Gaussian with mean and variance
\begin{align}
&\hat{\bm{\mu}}_{\bm{w}_{\hat{k}}}=\bm{\mu_{\bm{w}_{\hat{k}}}}+\bm{L} \big(\bm{y}^*_{t}-\bm{\Phi}_{t}^\top\bm{\mu_{\bm{w}_{\hat{k}}}} \big), \\
&\hat{\bm{\Sigma}}_{\bm{w}_{\hat{k}}}=\bm{\Sigma_{\bm{w}_{\hat{k}}}}-\bm{L}\bm{\Phi}_{t}^\top\bm{\Sigma_{\bm{w}_{\hat{k}}}},
\label{eq:munew}
\end{align}
where
\begin{equation}
\bm{L}=\bm{\Sigma}_{\bm{w}_{\hat{k}}}\bm{\Phi}_{t} \big(\bm{\Sigma}^*_y +\bm{\Phi}^\top_{t}\bm{\Sigma}_{\bm{w}_{\hat{k}}}\bm{\Phi}_{t} \big)^{-1}.
\label{eq:L}
\end{equation}
and ${\bm{w}_{\hat{k}}}, {\bm{\Sigma}_{\hat{k}}}$ are the weight and covariance of the identified $\hat{k}$ ProMP.
In situations where the motion capture system is highly accurate, we can consider $\bm{\Sigma}^*_{y}$ to approach zero, signifying that there is very little uncertainty in the observed data.

\subsection{Object Detection and Pose Estimation}
\label{sec:object}
In order to accurately perceive and interact with objects, the estimation of their pose is crucial. 
For object pose estimation, we utilized ArUco markers. This method serves as a temporary solution while the integration of state-of-the-art object detection algorithms in our framework is ongoing.
The user interface (UI) displays the estimated pose as an overlaying virtual object, generated from CAD data. 

In certain situations, manual override of object pose estimation may be necessary. We achieve this by utilizing the virtual objects in the interface (Section \ref{sec:interface}). The user can interact with these virtual objects and manually adjust their pose to match the real-world objects based on the point cloud from the depth camera and LIDAR on the robot.


\subsection{Affordance Templates (ATs)}
\label{sec:affordance}
An AT \cite{pettinger2020} is an adjustable pairing of 3D object geometries and sequence of robot actions represented in object-centric coordinates. At run-time, virtual object frames are registered by the user, defining waypoint frames that are converted into robot specific end-effector frames. To facilitate the creation and management of ATs, we have developed our user interface that allows for easy editing, saving, and loading.

\subsection{Blending ProMPs to ATs}
\label{sec:blending}
The generation of training trajectories for ProMPs does not require high precision. The object pose information ensures accuracy because ProMPs are updated to reach the detected object pose. However, the object's centroid may not align with the desired grasping point encoded in the AT. Hence, transitioning to the AT needs to occur before the ProMP motion is completed. To achieve a smooth transition, a policy blending mechanism adjusts the ProMP-generated motion $\bm{y}_{ProMP}$ by incorporating the first sample from the AT-generated trajectory $\bm{y}^{0}_{AT}$ . This process determines the refined reference for the movement using the following blending equation:
\begin{equation}
\bm{y}' = (1-\alpha)\bm{y}_{ProMP}+\alpha\bm{y}^{0}_{AT}.
\label{eq:blending}
\end{equation}
Here, $\alpha$ represents the blending coefficient, which is defined by the function:
\begin{equation}
\alpha(x)=\frac{1}{1+a ~ e^{-b(x-c)}},~ ~  \alpha\in~ [0,1] \\
\label{eq:alpha}
\end{equation}
The value of $x$ is given by $x =\frac{i}{N_b}$, where $i$ ranges from $0$ to $N_b$, and $N_b$ is the number of blending samples. The values of $a, b,$ and $c\in \mathbb{R}$ are chosen to ensure that $\alpha(0)=0$, $\alpha(1)=1$, and $0.8<\alpha(0.7)<0.9$. This allows for an early transition to the affordance, discarding any imprecision present in the final part of the recorded training trajectories. 

\subsection{Robot Controller}
\label{sec:control}
The robot uses a momentum-based whole-body controller that is framed as a quadratic program (QP) \cite{koolen2016}. The controller's primary task is to track a desired rate of change of momentum, but it can simultaneously track a set of external motion objectives for the robot's pelvis height, chest orientation and arm configurations. By solving the QP, the controller produces a joint acceleration vector and contact wrenches. These values are then used to calculate the desired actuator torques through inverse dynamics. 
In the context of humanoid teleoperation \cite{kourosh2021survey}, references for the tracked body segments are determined using an Inverse Kinematics (IK) QP. This IK QP ensures the feasibility and safety of the specified references \cite{jorgensen2019}. The output of the Inverse Kinematics QP is then supplied to the momentum-based whole-body controller, further enhancing the overall performance and balance of the robot's movements during teleoperation and assistive autonomy scenarios. For a more in-depth understanding of the whole-body controller, additional detailed information can be found in \cite{koolen2016}.
\begin{figure}[!t]
\centering
\includegraphics[width=\linewidth]{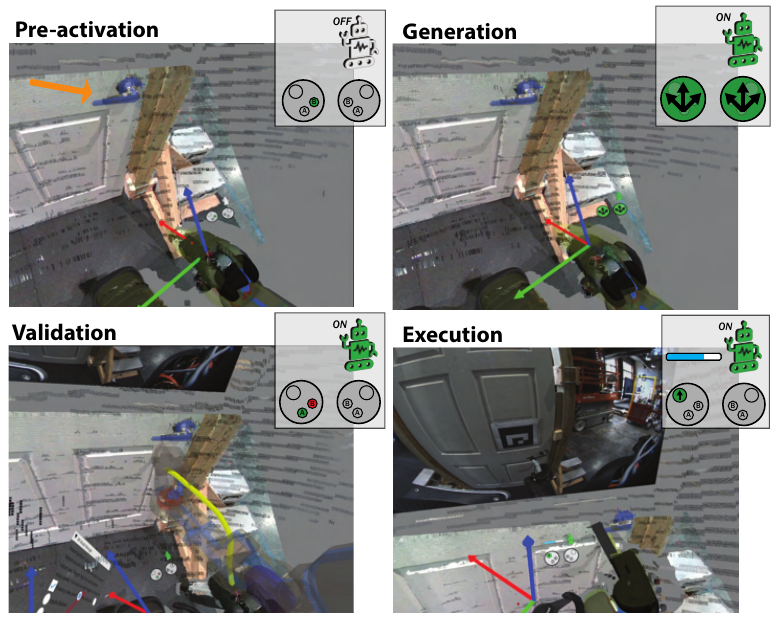}
\caption{Operator's view in mixed reality while remotely controlling the robot using the proposed teleoperation assistance. The virtual panel provides real-time guidance, presenting the user with essential instructions for each step of the operation. \textbf{Pre-activation.} An object has been detected (highlighted in blue and with an orange arrow) for which the autonomy is available, and the user can activate the teleoperation assistance.  \textbf{Generation.} The user can start doing the task and provide initial input to the autonomy that will adapt accordingly. \textbf{Validation.} The user can preview the proposed motion via a ghost robot and spline trajectories, validate it or reject it. \textbf{Execution.} The user can execute the proposed motion via joystick control.}
\label{fig:4stepsAndPreview}
\end{figure}

\begin{figure*}
\centering
\includegraphics[width=\linewidth]{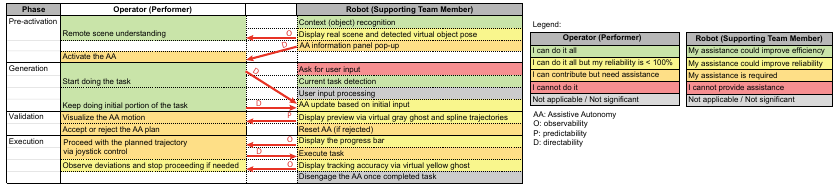}
\caption{Interdependence analysis of a teleoperated task using the assistive autonomy.}
\label{fig:IA}
\end{figure*}

\section{Mixed Reality Interface}
\label{sec:interface}
Throughout our work, we have formulated a set of robust principles that inform the design of human-robot operator interfaces. These principles have a central focus on three essential elements: observability, predictability, and directability \cite{johnson2014coactive}.
Essentially, the user should be able to observe the robot's current state and behaviors, predict its actions in response to instructions, and direct the robot's next actions.

Our current interface includes a camera view to display the robot's perspective, alongside a created ``digital twin'' world view. This world view integrates all perception inputs from a depth camera and LIDAR, as well as a third-person view of the robot generated from proprioception sensors and state estimation. This interface provides the user with the option to operate the robot using a mouse and keyboard or via a VR headset and controllers. The latter mode, which we call kinematics streaming, allows the operator to directly control the robot's hands (and other parts via VR trackers) and use the VR controller triggers to open and close the grippers. During kinematics streaming, a yellow virtual ghost robot appears, overlaying the robot visualizer model (Fig. \ref{fig:4stepsAndPreview}). This ghost robot serves as a reference, representing the solution found by our IK algorithm based on the current VR input.

Fig. \ref{fig:IA} depicts the interdependence analysis \cite{johnson2014coactive} highlighting the relationships between the user and the robot during a teleoperation task facilitated by the assistive autonomy and showing how Coactive Design principles govern their interaction. 
While the user is directly controlling the robot and whenever an object is detected for which the assistive autonomy is available, a virtual object representing the detected object appears in the UI at the estimated location.
An information panel also appears on top of the right VR controller (Fig. \ref{fig:4stepsAndPreview}). This panel informs the user that the assistive autonomy is available and can be activated by pressing a dedicated button on the left controller.

Once activated, the panel prompts the user to move and start performing the task (Fig. \ref{fig:4stepsAndPreview}). This allows the robot to analyze the user's initial inputs and determine trajectories for each controlled body part accordingly. At this stage, the robot can either be directly controlled by the user or remain in an idle state. The assistive autonomy processes the motion, identifies the task, and adapts to the user's preferred way of performing it. 
Following this initial processing, the system presents a preview of the computed motion to the user. Predictability over the robot actions is provided through a visualization of the robot's motion with a gray ghost preview and spline trajectories (Fig. \ref{fig:4stepsAndPreview}). At this point, the user has the option to either accept or reject the suggested motion by pressing a dedicated button on the VR controller. The preview of the planned trajectory determined by the assistive autonomy can be disabled if the user wants to speed up operation.

Upon validation, the user gains control over the robot and can direct the movements along the computed trajectories by tilting the joystick of a controller forward. This enables the user to control the robot's actions frame by frame. In fact, the motion can be paused by stopping the tilting motion at any time. To provide extra safety, a yellow ghost robot model is displayed as feedback on how well the robot is following the reference trajectories. If the operator observes significant deviations between the yellow ghost and the actual robot's movement, they can pause the motion, exit the assistive autonomy mode, or provide a different initial input. Thus, safety is guaranteed by the joystick control (directability) and the visualization of the yellow ghost (observability).

The information panel was designed with key usability heuristics in mind, such as visibility of system status, matching the system with the real world, and recognition over recall \cite{nielsen1994enhancing, adamides2014usability}. The assistive autonomy mode state is indicated by the color of a small robot icon (green for engaged, white for disengaged), and a horizontal progress bar shows the completion percentage of the task, helping the user observe robot's status (Fig. \ref{fig:4stepsAndPreview}). As the operator tilts the joystick forward, the blue bar expands, and when it reaches the edge, the task is complete, and the robot icon turns white. The panel utilizes representations of real-world VR controller icons for more intuitive control. Additionally, it guides the operator through the steps required to direct the assistive autonomy by highlighting the corresponding buttons or actions in green (Fig. \ref{fig:4stepsAndPreview}). 


\begin{figure}
\centering
\includegraphics[width=\linewidth]{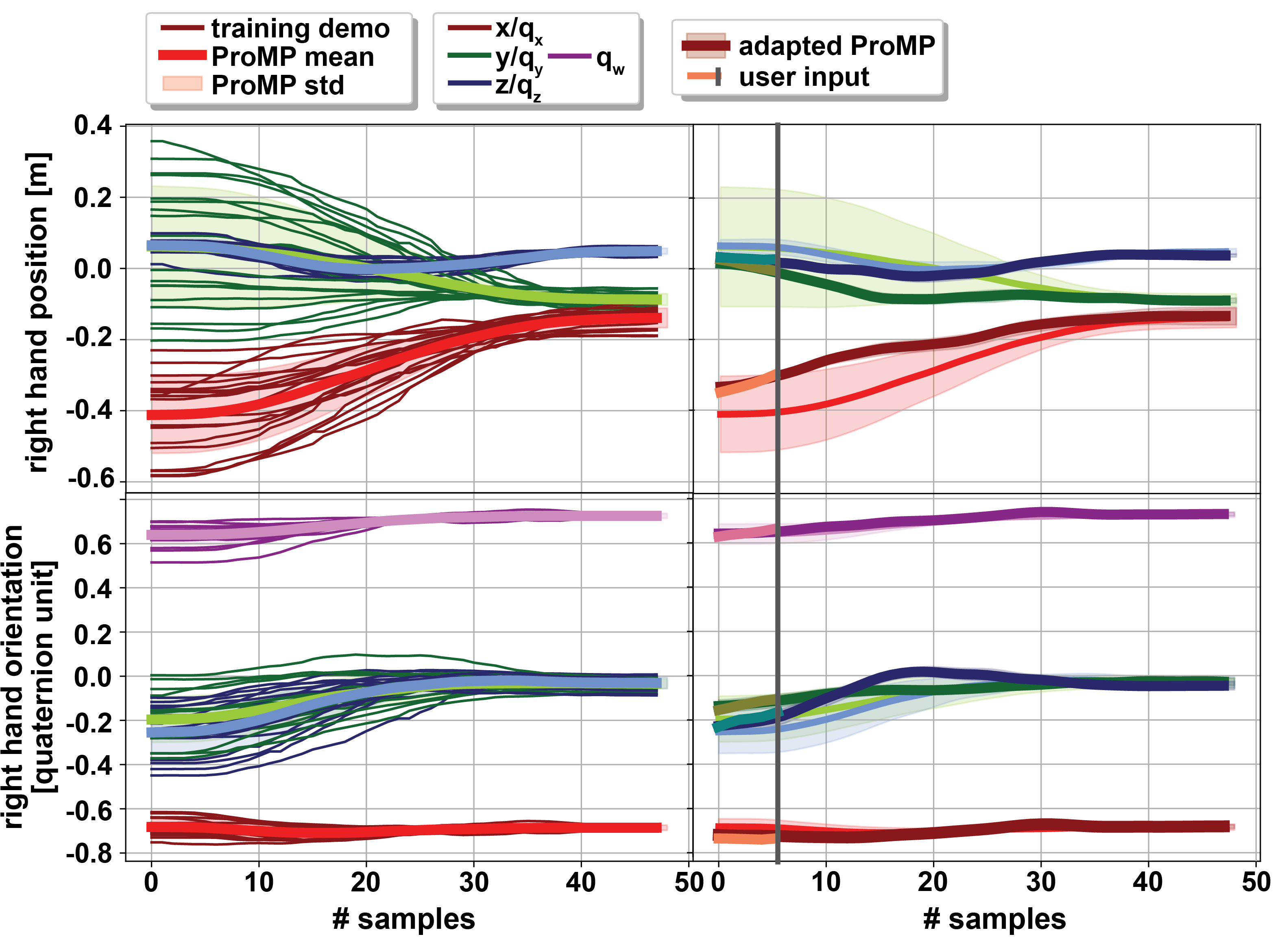}
\caption{ProMP-based reference trajectories for reaching the door handle.
     Left column: learned ProMPs (light thick lines with transparent regions) alongside the corresponding 20 demonstrations (dark thin lines), expressed in the door handle frame. These demonstrations include reaching motions from different approach locations. Right column: updated ProMPs (dark lines) after observing the user input (lighter short lines followed by a vertical gray bar).}
\label{fig:door}
\end{figure}
\begin{figure}
\centering
\includegraphics[width=\linewidth]{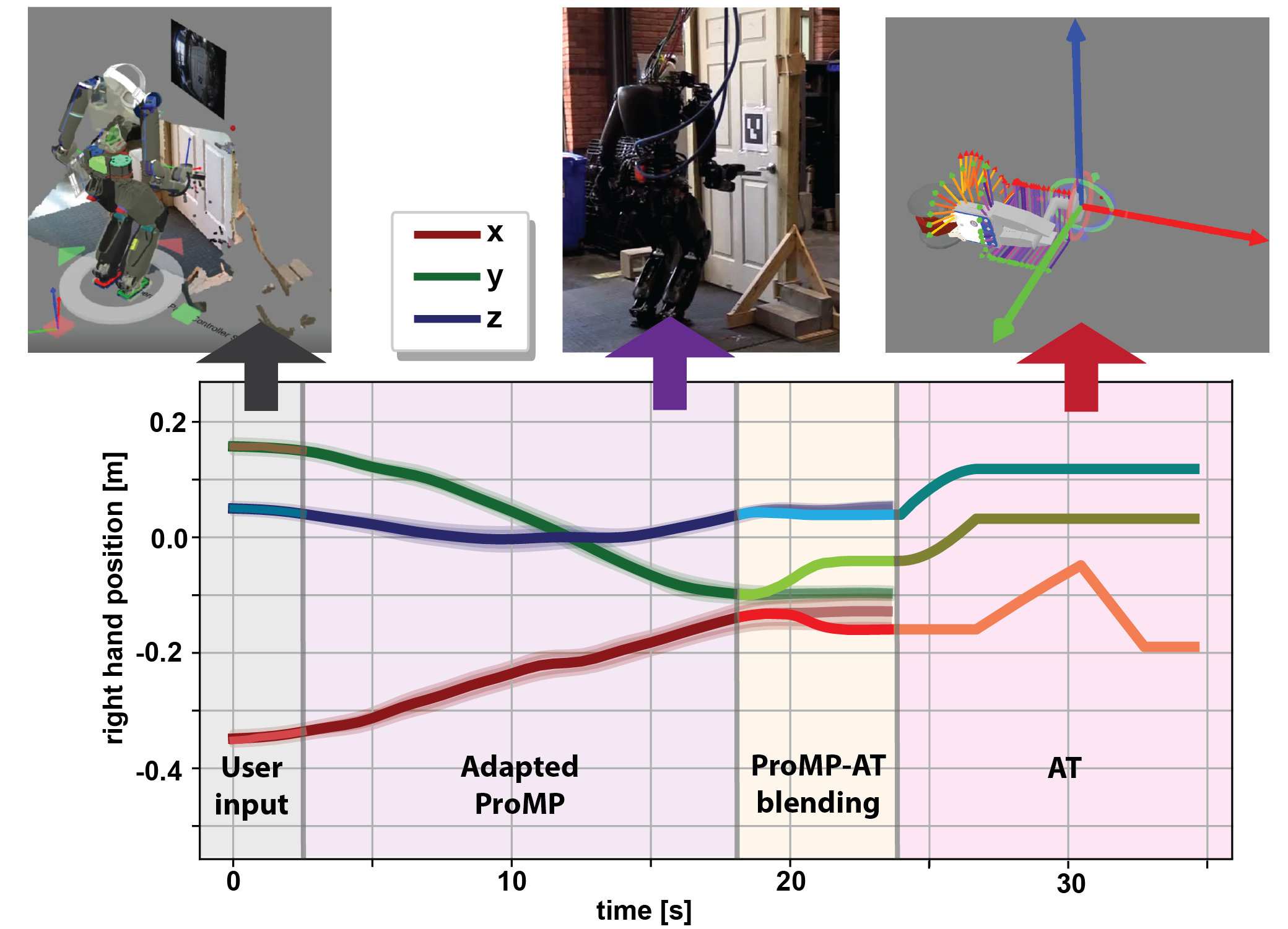}
\caption{ProMP-AT blending policy during teleoperation. The reference trajectories for the robot during the teleoperation process are illustrated. In the first part, the trajectories are the user's input with the robot following their motions. After the user has provided some input, they can stop. The ProMP trajectory is updated according the initial user input, allowing the robot to execute the updated motion. When the ProMP motion reaches its end, a blending mechanism smoothly transitions between the ProMP and the AT trajectories. Finally, the AT motion is executed.}
\label{fig:ProMP-AT}
\end{figure}

\section{Teleoperation Assistance in Action}
\label{sec:exp}

\subsection{Experiments}
\label{sec:setup}
We conducted experiments with Nadia\footnote{\url{https://boardwalkrobotics.com/Nadia.html}}, our advanced humanoid robot, provided with 31 degrees of freedom and partially powered by hydraulics. Users were equipped with a Valve Index VR headset and controllers to operate the robot. 
To evaluate the effectiveness of our approach, we conducted a door opening task. The training phase consists in having a single expert user perform a given task several times and in different ways. Hence for the door opening task, we collected data by attempting to reach the door handle from various angles and positions, using the VR controllers to control the robot's hand poses. This resulted in a diverse set of training data, which allows the system to learn and adapt to different scenarios that may be encountered during teleoperation, including variations in environmental conditions.

The training data is recorded in the door handle frame, which provides additional flexibility and adaptability to the detected pose of the door handle. This adaptive approach ensures that the robot can successfully complete the door opening task while compensating for variations in the environment and any inaccuracies in the user's input, especially when approaching the real object due to poor situational awareness. As shown in Fig. \ref{fig:door}, the updated ProMPs also align with the initial user's intention. This enables the user to effectively direct the robot's motion as they desire, while maintaining  fidelity to the demonstrations.
When approaching the object, a blending policy gradually shifts the input from the ProMP-calculated trajectory to the AT, as illustrated in Fig. \ref{fig:ProMP-AT}. This blending mechanism ensures smooth transitions between the autonomous motion generated by the ProMP and the controlled manipulation governed by the AT. By seamlessly integrating these two components, the robot can effectively and accurately interact with the object, reducing the risk of errors or unintended consequences during physical interaction.

To further emphasize the adaptability to the user intention of our system, we extended our evaluation to include a punching task. The unique aspect of this task lies in its multifaceted execution: a single action, such as a punch, can be manifested in diverse techniques, each so distinct that they're individually labeled, like hook punches, uppercuts, and jabs. Instead of devising separate ProMPs for each distinct punching technique and target, we opted for a more integrated strategy. A singular ProMP was developed to cover all techniques, adapting to different targets within the task. This approach aligns with the concept of having a limited number of movement primitives associated with each context.

\begin{figure*}
\centering
\includegraphics[width=17cm]{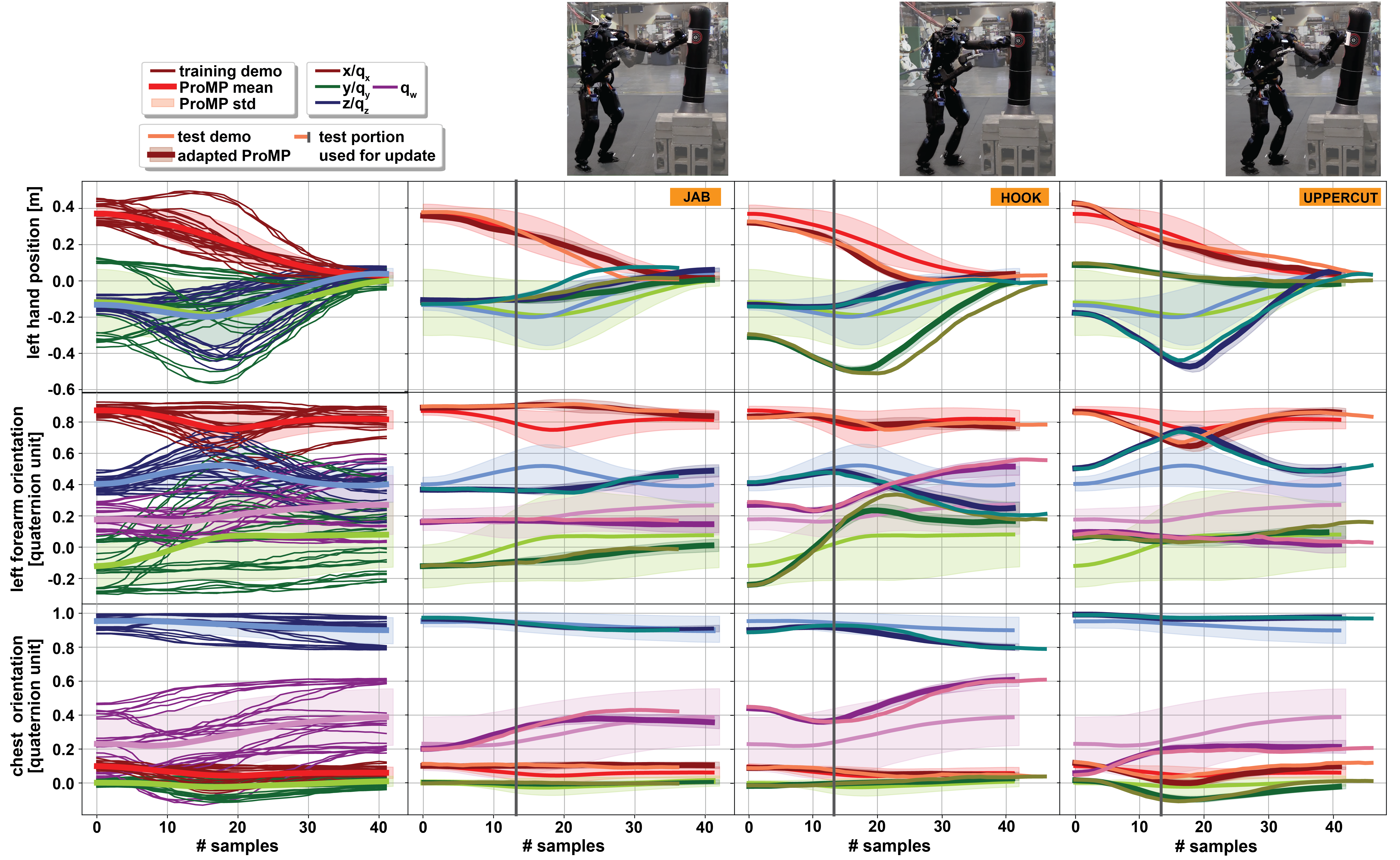}
\caption{ProMP adaptation during punching motion.
    The ProMPs for the most relevant body parts involved in a left punching motion are shown. Left column: learned ProMPs (light thick lines with transparent regions) alongside the corresponding 27 demonstrations (dark thin lines), expressed in the target frame. These demonstrations include 9 jabs, 9 hook punches, and 9 uppercuts. Remaining columns: updated ProMPs (dark thick lines) after observing a portion of the test trajectories (portion of the lighter thinner lines up to the vertical gray bar). We compare these updated ProMPs with the previously learned ProMPs. From left to right, the test trajectory represents a jab, a hook punch, and an uppercut, respectively.}
\label{fig:punching}
\end{figure*}

By employing this unified ProMP for the punching task, we aimed to assess the system's ability to generalize and adapt its punching actions to the different ways the user wants to perform the motion\footnote{Note that the punching task does not incorporate an AT; hence no blending is involved.}.
Differently from the door opening task, the punching motion required the coordination of multiple body parts, necessitating additional ProMPs for the forearm orientations, and chest orientation. In this case the user was also equipped with three Vive trackers 3.0 that tracked the forearms and chest motion. Fig. \ref{fig:punching} shows the ProMPs adaptation during a left punching motion. The left column displays the learned ProMPs alongside the corresponding demonstrations, while the remaining columns show the updated ProMPs after observing a portion of the test trajectories. The updated ProMPs align with the user's intended punching motion, allowing the robot to punch different targets with the user's chosen punching technique. 

Table \ref{tab:rms} reports the mean Root Mean Square (RMS) comparison -- with associated standard deviation ($SD$) -- between the trajectories generated by the user with those generated by the assistive autonomy after using a portion (about a third) of the same user directed trajectories as observation.  The mean is calculated based on 10 test trajectories for the punching motion\footnote{ The RMS error for a single test is calculated by evaluating the errors between each sample of the two compared trajectories. Next, the overall RMS in position or orientation is computed, followed by calculating the mean and $SD$ over all the tests.}. The RMS errors across different motion components indicate a remarkably small deviation between the trajectories generated by the assistive autonomy system and the user-directed robot motion (2.35$cm$ with a standard deviation of 0.87$cm$ for position and no more than 0.05$rad$ with a standard deviation of 0.02$rad$ for orientation). The similarity between the adapted trajectories and the user's intended motion demonstrates the system's ability to perform the task as intended by the user.
Videos and data of the experiments are available online\footnote{Dataset: \url{https://doi.org/10.5281/zenodo.8215964}, video: \url{https://youtu.be/oN-FD6YnF2c}}.

\begin{table}[h]
\caption{Assistive autonomy motion vs user directed robot motion for the left punching task.}
\begin{center}
\begin{tabular}{cc}
\hline 
&\multicolumn{1}{c}{\textbf{RMS error adapted ProMP vs test}}\\
\cline{1-2}
left hand position [cm] & 2.35 (0.87)  \\
left hand orientation [rad] & 0.03 (0.01)  \\
left forearm orientation [rad] & 0.05 (0.02)   \\
chest orientation [rad] & 0.04 (0.01) \\

\hline
\end{tabular}
\label{tab:rms}
\end{center}
\end{table}

\begin{table}[h]
\caption{Results from Feasibility Studies}
\begin{center}
\begin{tabular}{cccc}
\hline 
&\textbf{Completion Rate} &\textbf{Time on Task} &\textbf{Failed Attempts}\\
\textbf{Experts} &\textbf{[\%]} &\textbf{[s]} &\textbf{\#}\\
\cline{1-4}
DC & 75.0 (16.7) & 81.1 (7.1) & 1.9 (0.5)  \\
DC-AA & 100 (0.0) & 40.2 (11.7) & 0.0 (0.0) \\
\hline 
\textbf{Novices} & & &\\
\cline{1-4}
DC & 58.3 (31.9) & 86.6 (22.6) & 2.0 (0.7)  \\
DC-AA & 100 (0.0) & 43.3 (15.8) & 0.0 (0.0) \\
\hline

\multicolumn{4}{l}{DC: Direct Control, AA: Assistive Autonomy}
\end{tabular}
\label{tableResult}
\end{center}
\end{table}

\subsection{Feasibility Studies}
\label{sec:feasibilityStudy}

The IHMC Robotics Team conducted feasibility studies to test whether the humanoid robot can achieve the door opening task via teleoperation\footnote{The feasibility studies were classified as Non-Human-Subject Research (NHSR) and were conducted under approval from IHMC's IRB. No human research participants were involved or evaluated.}. A total of eight members of IHMC participated in the feasibility studies. The group of participants included four expert users who were well acquainted with the system and its operation, and four individuals who were not familiar with the system and that have declared not having used VR nor directly teleoperated a robot before. 
Each user was tasked with performing the door opening activity using two different methods: (i) direct control, and (ii) direct control with assistive autonomy. For each method, the users had to repeat the teleoperated task three times.

During each repetition, the robot's initial position and approach angle were varied relative to the door. Specifically, one trial was conducted with the robot positioned on the left side of the door, another at the center, and the third one on the right side of the door\footnote{\url{https://youtu.be/KeQwt9a9wV8}}. As a result, each user performed a total of six door opening trials - three for each of the robot positions, using each of the teleoperation methods. The following measures were collected: task completion rate, time on task, and the number of failed attempts. A trial time limit of 120 seconds was set, based on observations made during the development phase, and multiple door opening attempts were allowed within the time limit. Within the trial, we recorded the number of failed attempts as the number of times the robot was unable to successfully grasp the door handle, unlock the door after turning the handle, or both. Table \ref{tableResult} presents the summarized results of the feasibility studies, displaying the mean values with their corresponding $SD$.


\section{Limitations and Future Work}
\label{sec:limitations}
The system does not dynamically adjust to the user's pace in real-time. It is capable of registering the user's movement and can estimate or be pre-set to a custom speed. Nonetheless, it fails to respond to variations in the user's speed during live input, as showcased in the supplementary online video\footnote{\url{https://youtu.be/8EyQscc6PWc}}. This issue is particularly significant in the context of human-robot interactions and can be resolved by utilizing Interaction Primitives \cite{campbell2017}, which are an extension of ProMPs that facilitates real-time speed adaptation.

Another drawback of the current approach is that attempting to adapt the model based on observations that deviate significantly from the trained standard deviation can lead to improper conditioning of the model, resulting in undesirable behaviors$^6$. During the door-opening task within our feasibility study, such issues did not arise, as the demonstrations encompassed a wide range of possible methods and angles of approach. Nevertheless, in scenarios such as punching tasks, users might execute the action in unforeseen and alternative manners. To mitigate this in future developments, we plan to introduce a validation step for such trajectories, which will identify and rectify outliers by mapping them to the nearest point within the confines of the learned model.

In our future work, we also plan to conduct extensive user studies to thoroughly analyze the benefits of our approach in terms of user workload and situational awareness. This will provide valuable insights into how effectively the system meets user needs and expectations.

\section{Conclusions}
\label{sec:Conclusions}
Our work addresses the inefficiencies observed in humanoid robot teleoperation by introducing autonomous assistance features. We have identified challenges in traditional direct control methods and proposed a shared-control framework that combines user input as high-level guidance for the robot's autonomous capabilities. By leveraging ProMPs and ATs we facilitate robot operation. Further enriched by mixed reality technology, we provide users with an immersive platform that integrates the assistive autonomy. This unique framework equips users with interface elements designed for effective supervision, prediction, and management of the robot's assistance.


\bibliographystyle{IEEEtran}
\bibliography{IEEEabrv,biblio}

%



\end{document}